# Efficient Implementation of a Semantic-based Transfer Approach[1]

**Michael Dorna** and **Martin C. Emele**[2]

**Abstract.** This article gives an overview of a new semantic-based transfer approach developed and applied within the Verb*mobil* Machine Translation project [22]. We present the declarative transfer formalism and discuss its implementation. The results presented in this paper have been integrated successfully in the Verb*mobil* system.

## 1 Introduction

The application domain of the Verb*mobil* Machine Translation (MT) project [16, 22] is spontaneous spoken language in face-to-face dialogs. The scenario is restricted to the task of arranging business meetings, but the approach is intended to be extensible to other topics as well. The languages involved are English, German and Japanese. Apart from linguistic and cognitive research, this project is also a software-engineering challenge. The implementation of the Verb*mobil* Prototype consists of 25 software modules which are simultaneously under development.

This article presents a formalism and a implementation of a transfer approach based on proposals of [1, 6, 7, 8]. For a more linguistically motivated comparison of our approach with other MT[3] approaches cf. [10]. Our transfer approach transforms a semantic representation produced by various analysis steps of a source language into a semantic representation that is the input to generation for a target language. Therefore, the transfer equivalences abstract away from morphological and syntactic analyses of source and target languages. I.e., in general, these equivalences are described only on the basis of the semantics of the languages involved.

The work reported here combines and improves on the Shake-and-Bake approach of [3, 23] and the semantic approach of [2]. Instead of using sets of morpho-syntactic *lexical items (signs)*, as in Shake-and-Bake, we specify translation equivalences on sets of arbitrary *semantic entities*. Therefore, before entering the transfer component of our system, individual lexemes can already be decomposed or combined into sets of such entities, e.g. for stating semantic generalizations or providing suitable representations for inferences. Additionally to Shake-and-Bake and improvements like [8], we have extended the rules with restrictions describing local semantic or discourse contexts. These restrictions, which can also contain all kind of inferences, control the applicability of individual transfer rules.

Finally, we have filled the lack of control strategies for Shake-and-Bake-like approaches by developing a strategy for applying transfer rules and avoiding conflicts between applicable rules. These results can be applied to several other transfer-based approaches.

Our transfer rules do not, in themselves, trigger calls to the recursive application of subsequent transfer rules. This is the main distinction between our approach and the one presented in [2]. Furthermore, because the recursive rule application is not part of the rules themselves, our approach solves problems with discontinuous translation equivalences which the former approach cannot handle well.

The various semantic formalisms used in Verb*mobil* components [5, 12, 8] are variants of Underspecified Discourse Representation Structures (UDRS) [20]. These formalisms can be characterized by a semantic construction process using set union for composition and by the minimally recursive nature of the resulting representations. The underspecification used in the semantic representations supports the preservation of ambiguity, if the pragmatically most plausible equivalents contain the same kind of ambiguity in the source and target language. This is especially important in a practical machine translation system like Verb*mobil*. The disambiguation of different readings could require an arbitrary amount of reasoning on real-world knowledge and thus should be avoided whenever possible.

Together with other kinds of information, such as tense, aspect, prosody and morpho-syntax, the different semantic representations are mapped into a single representation called Verb*mobil* Interface Term (VIT) [4, 9]. This uniform data structure serves as input for semantic evaluation and for the transfer component. The transfer output is also a VIT which is used for generation.[4]

Section 2 of this paper sketches the semantic representations used in our semantic-based transfer approach. In section 3 we introduce the transfer formalism and give examples. In section 4 we describe our implementation and other procedural aspects of transfer. Finally, section 5 summarises the results.

## 2 Semantic Representations

In this section we characterise the semantic representations used in our transfer approach.

In Verb*mobil* variants of UDRS are used for semantic representations, cf. [5, 12, 8]. These semantic formalism share the ability to underspecify quantifier and operator scope together with certain lexical ambiguities. We assume an explicit event-based semantics [11, 19] with a Neo-Davidsonian representation of semantic argument relations. The Neo-Davidsonian notation is used for practical purposes such as accessing roles in transfer rules without looking at the predicates (verbs) introducing them.

---

[1] This work was funded by the German Federal Ministry of Education, Science, Research and Technology (BMBF) in the framework of the Verb*mobil* project under grant 01 IV 101 U. The work presented here is strongly influenced by discussions with our colleagues of the Verb*mobil* subproject TP 12 (Transfer). We would like to thank C. J. Rupp and the anonymous reviewers for their comments which helped to improve this article. The responsibility for the contents of this article lies with the authors.

[2] IMS, Universität Stuttgart, Azenbergstraße 12, D-70174 Stuttgart, {dorna,emele}@ims.uni-stuttgart.de

[3] For a detailed overview of different approaches to MT see e.g. [14], for transfer-based MT see e.g. [21, 18].

[4] See [17] for a description of the generator.

What is even more important is the flat representation of semantic entities as a set of labeled conditions. The labeling of semantic conditions is very useful since the recursive embedding of argument structure and operator scope etc. is no longer syntactically represented in a recursive representation, but achieved through the interpretation of additional labeling constraints. In this respect, labels act as pointers to the corresponding arguments.

Semantic representations encoded in feature structures,[5] e.g. like MRS, cf. [8], are open data structures with respect to unification. I.e. it is possible to add arbitrary information during transfer by applying rules. In contrast, our semantic representations are finite and restricted to the information given by the different linguistic analysis steps. We use non-recursive ground terms (with fixed arity) and finite sets represented by closed lists.

This set-oriented representation is essential for our transfer approach. The declarative part is easy to understand because set union and set difference are the only necessary operations. Access to individual semantic entities becomes simple and the application of individual transfer rules can be realized using set operations. Additionally, these operations can be optimized, e.g. by using ordered sets. Because the set elements are ground, element identification and access can be realized by matching and does not need the full power of unification.

Example (1a) shows one of the classical Verb*mobil* examples for rejecting a suggestion which can be translated as *that really doesn't suit me well*. The corresponding semantic representation is given in (1b).[6]

(1) a. *Das paßt echt schlecht bei mir.*
    *(lit.: that suits really bad for me)*
   b. `[l1:echt(l2), l2:schlecht(i1),`
      `l3:passen(i1), l3:arg3(i1,i2),`
      `l4:pron(i2), l5:arg2(i1,i3), l6:pron(i3)]`

Semantic entities in (1b) are represented as a Prolog list of labeled conditions. After the unification-based semantic construction, the logical variables for labels and markers, such as events, states and individuals, are skolemized with special constant symbols, e.g. `l1` for a label and `i1` for a state. Every condition is prefixed with a label serving as a unique identifier. Labels are also useful for grouping sets of conditions, e.g. for partitions which belong to the restriction of a quantifier or which are part of a specific sub-DRS. Additionally, all these special constants can be seen as pointers for adding or linking information within and between multiple levels of the VIT.

As already pointed out we use the VIT for input and output of transfer. Only the set of semantic conditions is shown in (1b); the other levels of the multi-dimensional VIT representation, which contain additional scope, pragmatic, morpho-syntactic and prosodic information, have been left out here. If necessary, such additional information can also be used in transfer for controling rule applicability, in semantic evaluation for resolving ambiguities and in generation for guiding the different lexicalization choices. Furthermore, it allows transfer to make fine-grained distinctions between alternatives in cases where the semantic descriptions of source and target language do not match up exactly.

Semantic operators like negation, modals or intensifier adverbials, such as *echt*, take extra label arguments for referring to other elements in the flat list which are in the relative scope of these operators.[7]

This form of semantic representation has several other advantages, too. Coindexation of labels and markers in the source and target parts of transfer rules ensures that the semantic entities are correctly related and hence obey semantic constraints which are not part of a rule itself, but possibly linked to it. I.e. sortal, topic/focus, etc. information may be preserved for lexical choices in generation. Because of decomposition semantic entities need not directly correspond to individual lexical items, e.g. in case of derivations and for a more fine-grained lexical semantics. With decomposition we can express generalizations and also apply transfer rules to parts of the decomposition.

## 3 Transfer Formalism

In this section we present our transfer formalism and discuss some examples.

### 3.1 General

The transfer process operates on a set representation of semantic entities for a source language (SL) and produces a different representation for a target language (TL). This process works on different data structures, i.e. no rewriting on one and the same set takes place.

The formalism provides modules. The splitting of name spaces and rules is important for composing different transfer steps. E.g. the front-end of our German-to-English transfer is a module used for postprocessing the semantic construction output, which means decomposition of nominalizations or combining several set elements to a general one. The composition has been shown very useful for modularization of different tasks into different modules.

Transfer equivalences are stated as relations between sets of SL semantic entities and sets of TL semantic entities. They are usually based on (parts of) semantic representations of individual lexemes, but might also involve representations of partial phrases for treating idioms and other collocations.

The variable free semantic representations allow the use of logical variables for labels and markers in transfer rules to express coindexation constraints between individual entities. Such entities are e.g. predicates, operators, quantifiers and (abstract) thematic roles. The skolemization prevents unwanted unification of labels and markers while matching individual transfer rules against the semantic representation.

### 3.2 Transfer Rules

The form of a transfer rule is given by

`SLSem, SLConds TauOp TLSem, TLConds.`

with a non-empty set of SL semantic entities (`SLSem`), an optional set of SL conditions (`SLConds`), an operator indicating the intended application direction (`TauOp`; one of `<->`, `->`, `<-`), a set of TL semantic entities (`TLSem`), and an optional set of TL conditions (`TLConds`). All sets are written as Prolog lists and optional conditions can be omitted.

We restrict the following discussion to the direction from German to English but most of the rules can be applied in the other direction as well.

(2) `[L:echt(I)] <-> [L:real(I)]`.

---

[5] Note that feature structures, which are by nature recursive data structures, can also be represented as (flat) sets of constraints, cf. [15]. Therefore our approach can be applied to all MT approaches based on this kind of data structure.

[6] For presentation purposes we have simplified the actual VIT representations.

[7] For the concrete example at hand, the relative scope has been fully resolved by using the explicit labels of other conditions. If the scope were underspecified, explicit subordination constraints would be used in a special scope slot of the VIT (see [5, 13] for subordination details).

**Natural Language Processing**          568          **M. Dorna and M. Emele**

The simple "lexical" transfer rule in (2) relates the German intensifier *echt* with the English *real*. The variables L and I ensure that the label and the argument of the German *echt* are assigned to the English predicate *real*, respectively.

(3) [L:passen(E),L:arg3(E,Y),L1:bei(E,X)] <->
    [L:suit(E), L:arg3(E,Y), L:arg2(E,X)].

The equivalence in (3) relates the German predicate *passen* with the English predicate *suit*. The rule not only identifies the event marker E, but unifies the instances X and Y of the relevant thematic roles. Despite the fact that the German *bei*-phrase is analyzed as an adjunct, it is treated exactly like the argument arg3 which is syntactically subcategorized. This rule shows how structural divergences can easily be handled within this approach.

(4) [L:passen(E), L1:bei(E,X)] <->
    [L:suit(E), L:arg2(E,X)].

The rule in (3) might be further abbreviated to (4) by leaving out the unmodified arg3. The transfer of arg3 can be handled by a metarule which passes on all semantic entities that are preserved between the source and target representation.[8]

### 3.3 Rule Restrictions

The rule in (5a) illustrates how an additional condition might be used to trigger a specific translation of *schlecht* as *not good* in the context of *passen*.

(5) a. [L:schlecht(E)], [L1:passen(E)] <->
       [L:neg(A), A:good(E)].
    b. [L:schlecht(E)] <-> [L:bad(E)].

The standard translation of *schlecht* as *bad* in (5b) is blocked for verbs like *passen (suit)*, that presuppose a positive attitude adverbial.[9]

One main advantage of having such conditions is the preservation of the modularity and compositionality of transfer equivalences. The transfer units remain small and, hence, the interdependencies between different rules are reduced.[10] The handling of such rule interactions is known to be one of the major problems in large MT systems.

A variation on example (1) is given in (6a).

(6) a. *Das paßt mir echt schlecht.*
    b. [L:passen(E)] <-> [L:suit(E)].

Here, the German verb *passen* takes an indirect object *mir* instead of the adjunct *bei*-phrase in (1). An appropriate transfer rule looks like (6b) which uses the metarule mentioned above for transferring the argument roles.

### 3.4 Overriding

In a translation system it is useful to be able to block or override transfer rules by stipulating exceptions. In a monotonic system without overriding it would be possible to apply the transfer rule in (6b) to the semantic representation in example (1b), which would not be appropriate. Whereas in the underlying rule application scheme assumed here, the more general rule in (6) will be blocked by the more specific rule in (4).

---

[8] The same effect could be achieved by adding one-to-one mappings for roles as well as other "interlingua" predicates.
[9] Our formalisms provides the definition of classes for transfer entities with similar properties instead of using specific lexical items like *passen* (see below).
[10] Note, that we have not used a (single) rule translating *schlecht passen* into *does not suit well* (or vice versa) because *schlecht* and *passen* are treated individually and the morpho-syntactic realization does not play a role in our approach. In the lexical Shake-and-Bake approach such a rule would have to be defined.

The specificity ordering is primarily defined in terms of the cardinality and by the subsumption order on terms. It also depends on the cardinality and complexity of conditions. For the *passen* example at hand, the number of predicates which are mentioned in a transfer rule defines the degree of specificity.

The conditional part of transfer rules can also be used for interaction with a domain representation, e.g. to get information about dialog acts or sortal restrictions.

(7) a. [L:termin(I)],[sort(I)=< ~temp_point] <->
       [L:date(I)].
    b. [H:termin(I)] <-> [H:appointment(I)].

The rule (7a) uses an additional condition which calls the external domain model for testing whether the sort assigned to I is not subsumed by the sort temp_point. Rule (7b) can serve as a kind of default with respect to the translation of *Termin*, in cases where no specific sort information on the marker I is available or the condition in rule (7a) fails.

### 3.5 Classes

Our transfer formalism supports definable classes which can be used for abstracting from specific lexical entries. This has been shown to be useful for minimizing the number of transfer rules, for separating lexicalization of individual languages from transfer tasks and for accessing contextual information for a set of different elements of a class.

(8) a. type(de,date_verbs,
          [absprechen,anbieten,festlegen,gefallen]).
    b. [L:termin(I)],
       [L1:date_verbs(E),L1:arg3(E,I)] ->
       [L:date(I)].

(8a) shows the definition of a class of German verbs. The de is the language module in which the class is defined, date_verbs is the class name and the list members are the elements of the class. The usage of the class in the condition of rule (8b) triggers the translation of *Termin* as *date*, if *Termin* appears in a collocation with one of the elements of date_verbs. E.g. *einen Termin anbieten* would be translated as *suggest a date*. Otherwise the rule in example (7b) can be used again, e.g. for translating *einen Termin haben* as *have an appointment*.

### 3.6 Scope Resolution

As has already been mentioned, a transfer rule condition can trigger inferences which are necessary for transfer. One such inference example is scope resolution.

(9) a. *Wir müssen noch ein Treffen vereinbaren.*
    b. *We still have to arrange a meeting.*
    c. *We have to arrange another meeting.*

The focus adverb *noch* in example (9a) can be translated as *still* (9b) or as *other* (9c) depending on the scope of *noch*. Because this difference is translation relevant the potentially underspecified focus scope has to be resolved, e.g. using prosodic or discourse information.

(10) a. [L:noch(F,S)] <-> [L:still(F,S)].
     b. [L:noch(F,S),L1:indef(I,R,S1)], [L1<F]
        <-> [L:another(I,R,S1), eq(S,S1)].

A "standard" one-to-one transfer rule for focusing adverbs is shown in (10a). F matches the focus label and S the scope label. The second rule (10b) maps *noch ein* to *another* and combines the scopes of the adverb and the indefinite article (quantifier) using eq.[11] The condition in (10b) checks whether the label of the quantifier is subordinated by the focus label of *noch*. In general, this check triggers on demand the scope resolution process which is completely hidden.

## 4 Implementation

Here we present our implementation and other procedural aspects.

The current transfer implementation consists of two parts: a) the Transfer Rule Compiler (TRC), which consists of a compiler together with a development and debugging interface, and b) the Transfer Runtime System (TRS), which consists mainly of runtime operations for the VIT and of process communication functions. The system was implemented in Prolog.

The TRC takes a set of rules like the one presented in section 3 and compiles them into an executable Prolog program. This program includes the selection of rules, the control of rule applications and calls to other processes if necessary.

The procedural interpretation of transfer equivalences can be described as a set of rewriting rules depending whether we translate from SL to TL or vice versa. To compute all possible translations in a naive way, we would have to construct the powerset of all possible partitions of the input set. We need to check for each subset of the partition whether there exists a matching rule. The resulting output would be obtained by joining the translations of all subsets of the partition. Subsets for which no matching translation rule exists cause the whole partition to be excluded from the set of possible translations. To select one of the possible translations we would need additional heuristics to choose the most appropriate translation. Such a naive implementation would require an exponential computation time even if there were only a single matching transfer rule for each subset.

In the actual Prolog implementation, sets are represented as Prolog lists, and set operations like membership can be computed by using standard list operations.

### 4.1 Indexing and Sorting

Because both the transfer input and the matching part of the rules (transfer units) are considered as sets of terms, the SLSems of the rules and the source semantic entities are sorted according to the default term order without loss of generality. Such a canonical set representation bears the following advantages:

1. We can exploit ordered set operations while searching for matching elements, e.g. we can use binary search instead of linear search.
2. We can put an index on transfer rules, namely by taking out the first element of a sorted SLSem. While mapping over the sorted input list of the SL, we only have to consider those rules whose index matches the current element. The Prolog built-in indexing facility ensures deterministic clause lookup (cf. next item).
3. We combine different rules with a common prefix into a single clause where different continuations are represented as local disjunctions. This reduces the search space drastically because we both reduce the number of computed clauses with respect to the rules defined and also only look for each prefix once. This affects neither the correctness nor the completeness.

### 4.2 Specificity Ordering

Our transfer approach deals easily with partial analyses as they are often the result of syntactic and semantic analysis of spontaneous spoken language. If more information is available, specific rules can be triggered. If there is less information, the more general rule will fire.

Control of the applicability of transfer rules is based on definable specificity and is not immediately part of a rule itself. It is defined according to the following criteria:

- if rule $r_i$'s SLSem is longer than rule $r_j$'s, prefer $r_i$ over $r_j$;
- if rule $r_i$'s SLSem is more instantiated than rule $r_j$'s, prefer $r_i$ over $r_j$;
- if rule $r_i$'s SLConds is longer than rule $r_j$'s, prefer $r_i$ over $r_j$.

It is easy to argue for the control strategy assumed here. If a specific rule exists, this can be assumed to be appropriate for a specific case where the less specific ones should not be applied. And such a treatment can never be part of a rule itself. In addition, it allows the treatment of more general (i.e. less specific) rules as default rules, in case the more specific ones are not applicable. Whereas in a strictly monotonic system, one would have to add the completion of all positive conditions (i.e. by computing a disjunction of negative conditions) to ensure that the default rule can only be applied if none of the other rules can be triggered.

It is also possible to extend our approach to a hierarchical representation of semantic entities in the lexicon, which is merely an extension of the specificity ordering in use. The ordering between rules is then also determined by the ordering of a term hierarchy.

### 4.3 Rule Application

We can sketch the traversal of rules at runtime as follows:
1. select (remove) the first term of the sorted input list;
2. find a rule whose index term matches the selected term;
3. try to remove the rest of the SLSem of this rule from the input list; if the complete SLSems matches,[12] go to 4.); else goto 2.);
4. check the SLConds (if any); if this is successful, add TLSem to the TL output and go to 1.); else goto 2.).[13]

### 4.4 Some Formal Properties

A transfer rule $r_i$ is given by a tuple $<$LHS, RHS$>$ where LHS is a nonempty ordered set, i.e. a list, and RHS is a set which is not necessarily ordered. We ignore the condition parts of rules in this section. The start state of a transfer process is given by the tuple $<$SL, $\emptyset$ $>$ where SL is an ordered set of semantic entities as already mentioned. A final state has the form $<\emptyset$, TL $>$. For a sequence of rule applications (derivation) we write

$$<\text{SL}, \emptyset> \stackrel{r_j}{\Longrightarrow} <\text{SL}_1, \text{TL}_1> \stackrel{r_k}{\Longrightarrow} \ldots \stackrel{r_n}{\Longrightarrow} <\emptyset, \text{TL}>$$

where $r_j, r_k, \ldots, r_n$ are the rules applied, respectively. A rule application for a rule $r_j$ is defined by

$$<\text{SL}_{i+1}, \text{TL}_{i+1}> := <\text{SL}_i \backslash \text{LHS}_j, \text{TL}_i \cup \text{RHS}_j>$$

where $\backslash$ is set difference[14] and $\cup$ is set union. The rule which can be applied is restricted by the set $R$

$$R_j = \{r_j \mid \textit{first}(\text{SL}_i) = \textit{first}(\text{LHS}_j)\}$$

where *first* computes the first element of a list and $=$ is matching.

The compiler reduces rules which are not unique, i.e. after compilation $\forall r_j, r_k \text{ LHS}_j = \text{LHS}_k \rightarrow \text{RHS}_j \neq \text{RHS}_k$ holds. The application order is computed such that a rule is checked at most once. This prevents spurious ambiguities caused by different rule application orderings resulting in the same output. Given the example rules:

---

[11] For presentation purposes we have simplified the decomposed semantic representation of *another*.

[12] We can break off the search for an element of the SLSem remainder if we know from the set ordering that it can no longer appear in the input list.

[13] The TLConds are currently not supported.

[14] The actual difference operation uses matching for variable instantiation in transfer rules.

$$r_1: \quad <\{a,b\},\{d\}>$$
$$r_2: \quad <\{c\},\{e\}>$$

two possible derivations exist for $<\{a,b,c\},\emptyset>$ which are

(1) $<\{a,b,c\},\emptyset> \overset{r_1}{\Longrightarrow} <\{c\},\{d\}> \overset{r_2}{\Longrightarrow} <\emptyset,\{d,e\}>$ and

(2) $<\{a,b,c\},\emptyset> \overset{r_2}{\Longrightarrow} <\{a,b\},\{e\}> \overset{r_1}{\Longrightarrow} <\emptyset,\{d,e\}>$.

Because of the computation of $R$, the combination of rules sharing common prefixes together with the specificity ordering of rules this problem disappears. Here, the only possible derivation is (1).

## 4.5 Development Aspects and Coverage

Raw versions of transfer rules can be generated automatically using bi-lingual dictionaries, lemmatizers for predicate names together with abstract semantic class descriptions for argument frames. The result is very close to a lexicalist approach. After this step a manual refinement has to be performed, especially for the definition of contextual restrictions which are attached to rules. This last step improves the quality of transfer output dramatically, as might be supposed.

The TRS is embedded in the incremental and parallel architecture of the Verb*mobil* Prototype. Interaction with external modules, e.g. the domain model and dialog module or other inference components, is handled by a set of abstract interface functions which may be called in the condition part of transfer rules. From the definition of a declarative rule it is not apparent whether such an abstract function works on the local context as defined by the current VIT-representation or whether it requires interaction with external modules. Therefore, the resolution of contextual conditions is completely hidden from the user.

The debugging of rules is guided and supported by compiler options, e.g. producing runtime statistics and reporting actions like matching failures and applied rules. The compiler is quick, e.g. the current number of about 1700 transfer rules for German-to-English are compiled in less than 25 seconds. The average transfer time for a 15 word sentence is about 30 milliseconds. Hence the implementation and development of transfer rules can be done very efficiently.

## 5 Summary

We have presented a declarative transfer formalism, which provides the implementation platform for a semantic-based transfer approach. This approach avoids many of the problems of former transfer and interlingua approaches and is well suited for purposes of MT.

By compiling the declarative transfer correspondences into an executable program we obtain an efficient transfer system. In our system different control and implementation schemes may be explored without having to change the declarative rule descriptions. In addition, we have described the underlying control scheme for the application of competing transfer rules. This kind of control is required in every MT system even though most systems do not mention this.

Future work will include the automatic acquisition of transfer rules from tagged bilingual corpora and domain specific dictionaries. Furthermore, the transfer formalism will support domain switches, abstract semantic construction operations and macros for parameterized rule schemata.